\pdfoutput=1

\documentclass[11pt]{article}

\usepackage{EMNLP2022}

\usepackage{times}
\usepackage{latexsym}
\usepackage{amsmath}
\usepackage{amssymb}
\usepackage{yhmath}
\usepackage{diagbox}
\usepackage{enumitem}
\usepackage{graphicx}
\usepackage{CJKutf8}
\usepackage{floatrow}
\usepackage{url}
\DeclareFloatVCode{afterfloat}{\vspace{-0.08in}}
\usepackage[label font=bf,labelformat=simple]{subfig}
\usepackage{caption}

\usepackage[T1]{fontenc}

\usepackage[utf8]{inputenc}

\usepackage{microtype}

\usepackage{inconsolata}

\title{Language Models Are Poor Learners of Directional Inference}

\author{{\bf Tianyi Li$^{\diamond}$} \quad {\bf Mohammad Javad Hosseini$^{\diamond}$\thanks{ \quad Now at Google Research.}} \quad {\bf Sabine Weber$^{\diamond}$} \quad {\bf Mark Steedman$^{\diamond}$} \\
        $^\diamond$ School of Informatics, University of Edinburgh \\
        \texttt{\{tianyi.li, javad.hosseini\}@ed.ac.uk} \\
    \texttt{s.weber@sms.ed.ac.uk, steedman@inf.ed.ac.uk}
}

\begin{document}
\begin{CJK*}{UTF8}{gbsn}
\maketitle
\begin{abstract}
We examine LMs' competence of directional predicate entailments by supervised fine-tuning with prompts. Our analysis shows that contrary to their apparent success on standard NLI, LMs show limited ability to learn such directional inference; moreover, existing datasets fail to test directionality, and/or are infested by artefacts that can be learnt as proxy for entailments, yielding over-optimistic results. In response, we present BoOQA (Boolean Open QA), a robust multi-lingual evaluation benchmark for directional predicate entailments, extrinsic to existing training sets. On BoOQA, we establish baselines and show evidence of existing LM-prompting models being incompetent directional entailment learners, in contrast to entailment graphs, however limited by sparsity.
\end{abstract}

\section{Introduction}
\label{Sec:intro}

Pre-trained language models have shown impressive performance in natural language understanding (NLU), where prompting methods are widely used for fine-tuning. \cite{raffel_exploring_2020,NEURIPS2020_1457c0d6,schick_its_2021}

In this paper, we specifically investigate predicate entailment detection, an important sub-task of NLU and specifically, NLI. The task is to predict, given that predicate \textcolor{purple}{\textbf{p}} holds between arguments <\textcolor{blue}{\textbf{a}}, \textcolor{olive}{\textbf{b}}>, whether it can be inferred that predicate \textcolor{red}{\textbf{q}} also holds between <\textcolor{blue}{\textbf{a}}, \textcolor{olive}{\textbf{b}}>.
For instance, ``\textcolor{blue}{John} \textcolor{purple}{shopped in} \textcolor{olive}{\sc ikea}'' entails ``\textcolor{blue}{John} \textcolor{red}{went to} \textcolor{olive}{\sc ikea}'', but not ``\textcolor{blue}{John} \textcolor{red}{drove to} \textcolor{olive}{\sc ikea}''.

The primary distinction between predicate entailments and semantic similarity, apart from being focused on predicates, is that the former involves \textbf{directional} entailments as well as \textbf{symmetric} ones. Directional entailments are those $p \vDash q$ ($p$ entails $q$) where $q \nvDash p$; conversely, symmetric entailments are those  $p \vDash q$ where $q \vDash p$ as well (namely $p \equiv q$). 

Directional entailments are important for question answering, since they help filter out the spurious connections from knowledge sources to questions: knowing that \textcolor{blue}{John} \textcolor{purple}{went to} \textcolor{olive}{\sc ikea}, it is unsafe to infer that \textcolor{blue}{he} \textcolor{red}{shopped in} \textcolor{olive}{\sc ikea}, as he may have been there for other reasons. By symmetric similarity (i.e. paraphrase), the two events would be considered related, so a spurious inference chain would emerge; by directional entailments, it would be concluded that while the two events are related, the entailment holds only in the reverse direction, so the spurious connection would be avoided.

Current LM-prompting methods have reported positive results on predicate entailment detection \cite{schmitt_language_2021,schmitt_continuous_2021}. Since the masked-language-modelling objective naturally enables LMs to separate related and unrelated tokens, they are expected to be good paraphrase detectors; on the other hand, it is less clear whether they also distinguish the directionality of entailments.

To answer this question, we adapt the SOTA LM-prompting model \cite{schmitt_language_2021} as a gauge for the competence of its LMs, in particular RoBERTa \cite{liu_roberta_2019} and BERT \cite{devlin_bert_2019}. We apply it to various subsets of the common benchmark LevyHolt \cite{levy_annotating_2016,holt_probabilistic_2019}. We find that while it scores highly on the directional subset by \citet{holt_probabilistic_2019}, it otherwise shows poor ability in learning the directionality of predicate entailments. We find instead that the LevyHolt directional subset is infested with artefacts, to which LMs are overfitting.

These observations show that we need a robust evaluation benchmark for directional predicate entailments, independent of training sets. Inspired by \citet{mckenna_multivalent_2021} and \citet{chen_reading_2017}, we present BoOQA, a Boolean Open QA dataset in English and Chinese, which is closer to applications, adversarial to artefacts in supervision, and demands sensitivity to the directionality of entailments.

On BoOQA, we re-examine supervised and unsupervised LM methods along with various discrete entailment graphs (EG) \cite{hosseini_learning_2018,hosseini_open-domain_2021,chen_entailment_2022,li_cross-lingual_2022}. We find that the performances of supervised LM-prompting methods are indifferent to directional supervision, and are generally less competitive than suggested on LevyHolt; on the other hand, EGs reach decent precisions with their strongest edges, but are hit by sparsity and noisy unsupervised signals.

Our contributions can be summarized as follows: 1) We show that LevyHolt, the common directional predicate entailments benchmark, is infested by artefacts, allowing supervised methods to perform well by overfitting; 2) We verify that LMs, with supervised fine-tuning, show limited ability to learn directional entailments; 3) We present BoOQA, a robust, extrinsic, multilingual evaluation benchmark for directional predicate entailments, where various baselines are provided and analysed.\footnote{Our code and datasets are published at \url{https://github.com/Teddy-Li/LM-DirctionalInference}.}

\section{Background and Setup}
\label{Sec:background}

Language models have been used under a pretrain-finetune paradigm: the semantics of a token in context are learnt during pre-training and reflected in the dense encodings; when fine-tuning with a task-specific dataset, the model learns which area of its encoding space to look at.
Therefore, if a pre-trained LM \textbf{cannot} be fine-tuned
to solve a task, we cannot reject the null hypothesis that it does not encode the task. In \S \ref{Sec:prompting}, we look into RoBERTa \cite{liu_roberta_2019} and BERT \cite{devlin_bert_2019} in particular, and examine whether they can be fine-tuned to learn directional predicate entailments.

\paragraph{Model}  We adapt the supervised SOTA \cite{schmitt_language_2021}, a prompt fine-tuning method, for examining LMs.\footnote{There is a follow-up work \cite{schmitt_continuous_2021} to this, but we found it to have inferior generalisation performance; see Appendix \ref{supp:ss_probe} for details and a brief introduction.} We call it S\&S here and below. S\&S fits each premise-hypothesis pair into a few natural language prompts, such as ``\textit{\textcolor{blue}{John} \textcolor{purple}{shopped in} \textcolor{olive}{\sc ikea}, which means that \textcolor{blue}{John} \textcolor{red}{went to} \textcolor{olive}{\sc ikea}}''; they then convert the task into sentence classification over instantiated prompts. It is a simple SOTA with few additional parameters, and the architecture allows directional judgements. Thus, it is an ideal ``gauge'' for directional ability of LMs.

\paragraph{Dataset} So far there are two popular predicate entailment datasets: LevyHolt \cite{levy_annotating_2016,holt_probabilistic_2019} and Sherliic \cite{schmitt_sherliic_2019}. We use LevyHolt in our \S \ref{Sec:prompting} experiments, as it contains data entries ($p \vDash q$?) with their converses ($q \vDash p$?), making the ground truth directionality annotations available. We use the train/dev/test split as in \citet{schmitt_language_2021}.\footnote{Except when an entry and its converse appear in different splits (e.g. one in train, the other in dev), where we randomly assign both into the same split, so as to avoid information leakage.} In each data split, We further classify the entries into the following 4 sub-groups, in parentheses are the sizes of each sub-group in each split:
\begin{itemize}[itemsep=0pt]
    \item \textit{DirTrue} (251 / 64 / 892): directional true entailments where the premise entails the hypothesis, but not vice versa; for instance, \textit{\textcolor{blue}{Person} \textcolor{purple}{shopped in} \textcolor{olive}{Location} $\vDash$ \textcolor{blue}{Person} \textcolor{red}{went to} \textcolor{olive}{Location}};
    \item \textit{DirFalse} (251 / 64 / 892): directional non-entailments where the hypothesis entails the premise, but not vice versa; for instance, \textit{\textcolor{blue}{Person} \textcolor{purple}{went to} \textcolor{olive}{Location} $\nvDash$ \textcolor{blue}{Person} \textcolor{red}{shopped in} \textcolor{olive}{Location}};
    \item \textit{Paraphrases} (615 / 155 / 1939): symmetric paraphrases where the premise and the hypothesis entail each other; for instance, \textit{\textcolor{blue}{Person} \textcolor{purple}{arrived at} \textcolor{olive}{Location} $\equiv$ \textcolor{blue}{Person} \textcolor{red}{got to} \textcolor{olive}{Location}};
    \item \textit{Unrelated} (3255 / 831 / 9198): unrelated predicate pairs where the premise and the hypothesis have no entailment relations; for instance, \textit{\textcolor{blue}{Person} \textcolor{purple}{shopped in} \textcolor{olive}{Location} $\nvDash$ \textcolor{blue}{Person} \textcolor{red}{fell ill in} \textcolor{olive}{Location}}.
\end{itemize}


We define various subsets with pairs of sub-groups, which we introduce and discuss in \S \ref{Sec:prompting}.


\paragraph{Evaluation Metric} In predicate entailment detection, Area-Under-the-Curve with precision $>50\%$ ($\it AUC_{50\%}$) has been the metric in use \cite{hosseini_learning_2018,schmitt_language_2021}. It is a solid metric for comparison on the same dataset; however, we are comparing between different subsets, each with a different random baseline precision (i.e. the ratio of true entailments). If we were to set a common precision lower-bound, we would be biased toward those datasets with higher random baseline precisions. To make performance on different datasets comparable, we propose the metric of \textbf{Normalized AUC} ($\it AUC_{norm}$):

\begin{equation}
    AUC_{norm} = \frac{AUC_{\xi}-\xi}{1-\xi}
\end{equation}

where $\xi$ is the random baseline precision. Intuitively, $\it AUC_{norm}$ measures the ratio of area-above-random ($1-\xi$) that falls below the precision-recall curve ($\it AUC_{\xi}-\xi$), see Appendix \ref{supp:auc_norm_demo} for graphic illustration. $\it AUC_{norm}$ allows us to take into account all gains against the random baseline, and level performance on all datasets to the same scale. 

\section{Prompt Fine-tuning LM with LevyHolt}
\label{Sec:prompting}

In this section, we test for LMs' ability to learn directional entailments with the S\&S prompt-based gauge model. We use RoBERTa-base as the primary subject, as it is used by SOTA \citet{schmitt_language_2021}, and is sufficiently lightweight for experiments to run efficiently. In Appendix \ref{supp:lh_result}, we also report results on RoBERTa-large and BERT models for key experiments, where results are consistent. We use $S\&S_{subset}$ to denote S\&S model fine-tuned on each $subset$.

\begin{figure}[t]
  \centering
  \includegraphics[width=0.89\linewidth]{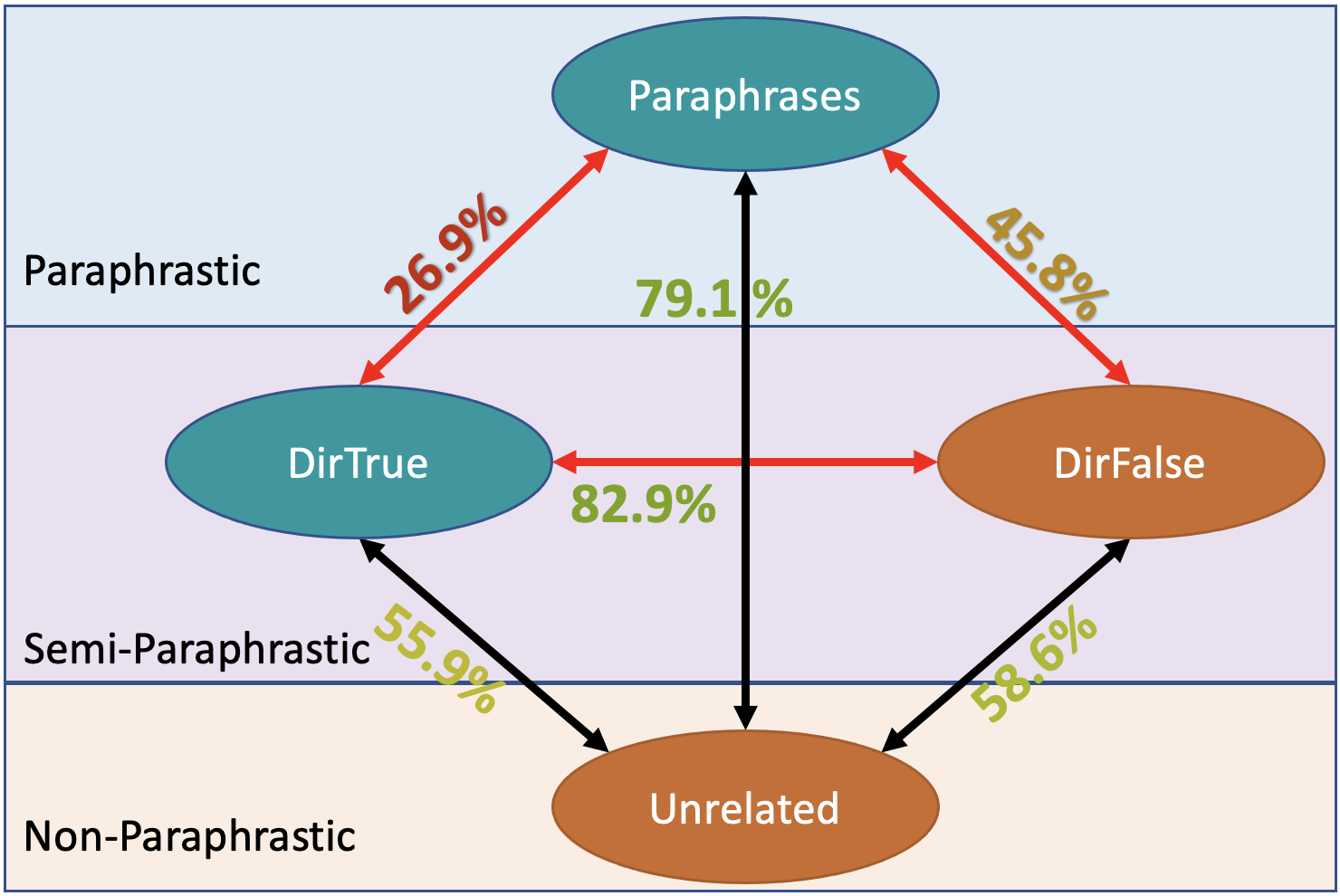}
  \caption{S\&S models on mesh of pairs of sub-groups, results in $\it AUC_{norm}$.}
  \label{fig:prompting:primary}
\end{figure}

Experiments are graphically summarized in Figure \ref{fig:prompting:primary}. Each edge denotes a LevyHolt subset made of the two sub-groups; the number on each edge is the $\it AUC_{norm}$ that S\&S achieves on each subset. For separating an \textcolor{teal}{Entailed} sub-group from a \textcolor{orange}{Non-entailed} one, the original labels are used; for separating two \textcolor{teal}{Entailed} or two \textcolor{orange}{Non-entailed} sub-groups, the one with more similar predicates (more paraphrastic) is assigned label ``1'', the other ``0''.\footnote{We acknowledge that \citet{schmitt_language_2021} use hand-crafted prompts tuned for entailment detection, so they may be sub-optimal for separating same-label sub-groups; we argue that fixed-prompt LM tuning models are not too sensitive to their specific prompts \cite{logan_iv_cutting_2021,webson_prompt-based_2022}; nonetheless, we also report results from a continuous-prompt model \cite{schmitt_continuous_2021} in Appendix \ref{supp:ss_probe}, where results are very similar.}

Note that we fit S\&S to a number of different subsets, so we cannot simply re-use the original hyper-parameters. Instead, to provide a level playing field, we follow \citet{schmitt_language_2021} in log-uniformly sampling 100 hyper-parameter sets from their specified ranges; for each subset, we choose the one with best dev set result.

If a method is insensitive to directional entailments, then it would treat entailments as similarity between unordered pairs of predicates; it would model \textit{Paraphrases}, \textit{DirTrue} and \textit{DirFalse} similarly, where \textit{DirTrue} and \textit{DirFalse} entries are conceptually somewhat ``semi-paraphrastic''.

If a method is sensitive to directional entailments, it should be able to discriminate between each pair of the four sub-groups. Particularly, it should additionally be able to separate sub-groups in the upper triangle of the mesh, coloured \textcolor{red}{red} in Figure \ref{fig:prompting:primary}. Among these three tests of directionality, \textit{DirTrue-DirFalse} is the most challenging: a method with no sensitivity to directionality should be at chance and get 0\% $\it AUC_{norm}$; this is traditionally called the \textit{directional subset} \cite{holt_probabilistic_2019}. For the other two subsets, a symmetric measure would do above random by identifying entries in \textit{DirTrue} / \textit{DirFalse} as statistically less similar than \textit{Paraphrases}.

Below we discuss findings around the mesh and the triangle.
The S\&S model yields 77.7\% $\it AUC_{norm}$ when trained and tested on full LevyHolt, which we provide for readers' reference.

\begin{figure}[t]
  \centering
  \includegraphics[width=0.89\linewidth]{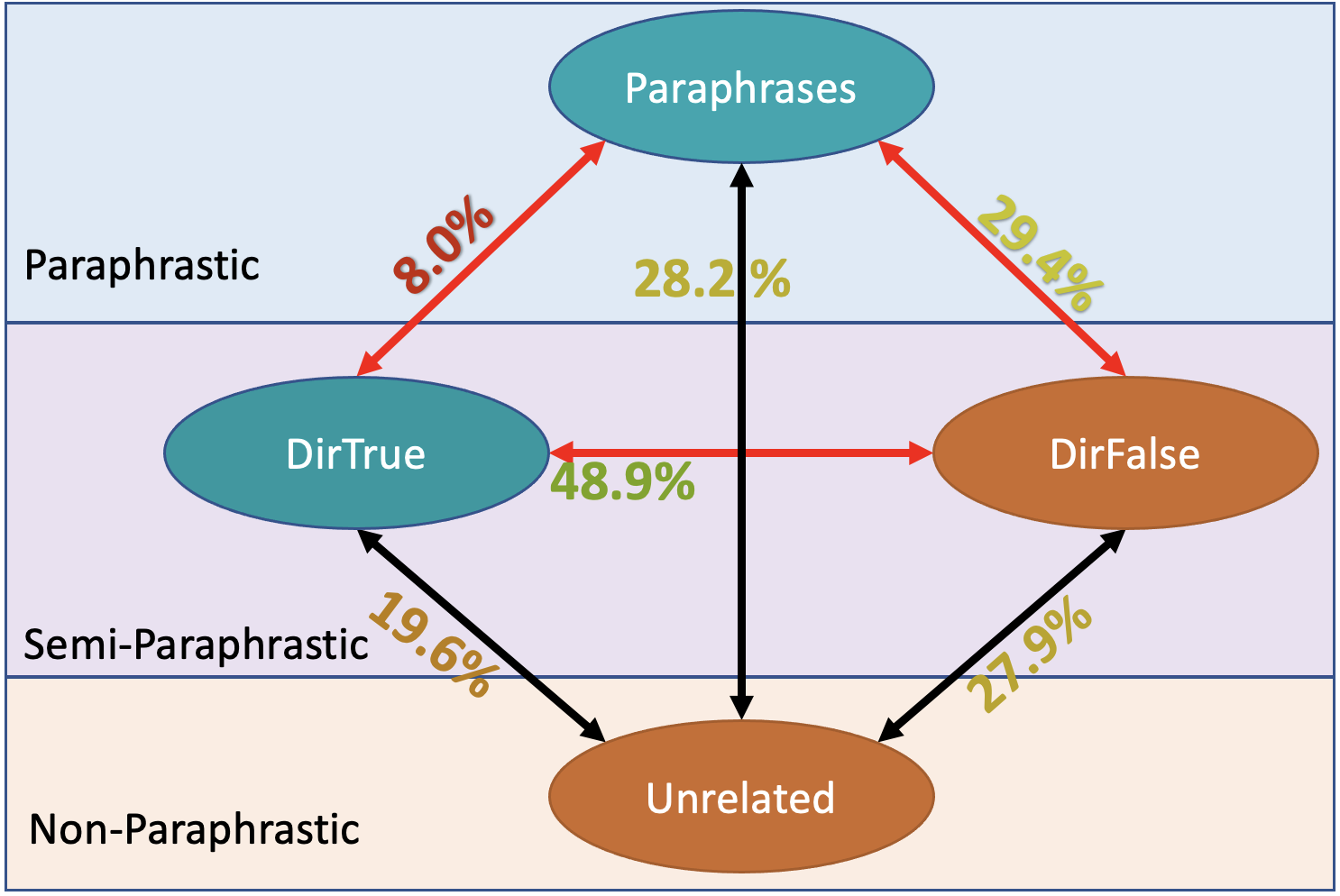}
  \caption{Hypothesis-only artefact baselines on mesh of pairs of sub-groups, results in $\it AUC_{norm}$.}
  \label{fig:prompting:hypoonly}
\end{figure}

\subsection{The S\&S Triangle}
\label{Sec:prompting:ss_triangle}

The red triangle in Figure \ref{fig:prompting:primary} presents mixed messages about the directionality of RoBERTa LM: on the most challenging \textit{DirTrue-DirFalse} subset, it achieves an apparently excellent $\it AUC_{norm}$ of 82.9\%; however, on the other subsets, it gets mediocre results at 26.9\% and 45.8\% respectively.

\begin{table}[t]
    \centering
    \begin{tabular}{|c|c|c|c|}
        \hline
        \diagbox[width=2.3cm, height=1.4cm]{Train/Dev}{Test} & Directional & Symmetric & Full \\\hline
        Directional & \textbf{82.9} & 9.9 & 24.7 \\\hline
        Symmetric & 0.2 & \textbf{79.1} & 61.3 \\\hline
        Full & 46.4 & 84.8 & \textbf{77.7} \\\hline
    \end{tabular}
    \caption{Generalization performance of RoBERTa-base S\&S classifier on the \textit{Directional} and \textit{Symmetric} subsets of LevyHolt. Values are in \% of $\it AUC_{norm}$.}
    \label{Tab:dir_and_sym}
\end{table}

\begin{table}[t]
    \centering
    \begin{tabular}{|c|c|c|}
    \hline
    \rule{0pt}{14pt} $\it AUC_{norm}$ (\%) & $S\&S$ & $\widehat{S\&S}$ \\\hline
    Para-DirTrue & 26.9 & 19.8 (-7.1) \\
    Para-DirFalse & 45.8 & 35.6 (-10.2) \\\hline
    \end{tabular}
    \caption{Comparison between $S\&S$ with regular and $\widehat{S\&S}$ with symmetric prompts. \textit{Paraphrases-DirTrue} and \textit{Paraphrases-DirFalse} subsets are concerned.}
    \label{Tab:symp}
\end{table}

For the \textit{directional} subset (\textit{DirTrue-DirFalse}), the 82.9\% $\it AUC_{norm}$ not only surpasses
the 77.7\% for Full LevyHolt, but is also on par with the 79.1\% for its mirroring \textit{Symmetric} subset (\textit{Paraphrases-Unrelated}), which should be easier by human judgement. Paradoxically for such a challenging subset, the \textit{Directional subset} enjoys the best performance in the mesh.

To understand this result, we did a generalisation experiment between \textit{Directional} and \textit{Symmetric}, the two disjoint, complementary subsets of LevyHolt. As results in Table \ref{Tab:dir_and_sym} show, classifiers from the two subsets do not generalise to each other, and neither does $S\&S_{Directional}$ generalise to full LevyHolt. That is to say, either the two kinds of ``entailments'', \textit{Directional} and \textit{Symmetric}, are different tasks from the LM's perspective, or the S\&S classifier is overfitting to the \textit{directional} subset.

For the \textit{Paraphrases-DirX} subsets, results are far less impressive. For reference, we train two strictly-symmetric S\&S models, one on \textit{Paraphrases-DirTrue}, the other on \textit{Paraphrases-DirFalse}. For these strictly-symmetric models we enforce all prompts to be in pairs of reverses (e.g. for the example in \S \ref{Sec:background}, we would add ``\textit{\textcolor{blue}{John} \textcolor{purple}{went to} \textcolor{olive}{\sc ikea}, which means that \textcolor{blue}{John} \textcolor{red}{shopped in} \textcolor{olive}{\sc ikea}}''). That way we guarantee from the input that no directional judgements can be made. We call these symmetric-prompt models $\widehat{S\&S}$. From the results in Table \ref{Tab:symp}, we find that for both \textit{Paraphrases-DirTrue} and \textit{Paraphrases-DirFalse}, there is only a modest difference between the performance of $S\&S$ classifier and $\widehat{S\&S}$. This shows that despite the results from the \textit{Directional Subset},
RoBERTa LM shows limited ability to detect directional entailments.




\subsection{The Artefacts Triangle}
\label{Sec:prompting:artefacts}

From previous discussions, we notice that the scores for
the \textit{directional} subset is anomalously high. Below we reveal that this anomaly is an effect of dataset artefacts, and that the artefacts in question are quite specific to the \textit{directional} subset and generally less prominent in the other subsets. Artefacts aside, the S\&S classifiers do not show strong abilities to learn directional entailments. 

It is difficult to identify sources of artefacts by manual inspection; on the other hand, \citet{poliak_hypothesis_2018} have shown that hypothesis-only (H-only) models can be a strong proxy for an artefact baseline. Inspired by their findings, we instead use H-only model as a proxy for the aggregated strength of artefacts. For H-only model we use a restricted version of S\&S classifier, where we mask all premises with the word ``true''.\footnote{We use ``\textit{true}'' to mask the premise because, when the premise is always true, the correctness of instantiated prompts depends solely on the hypothesis.}

We report the H-only model's results on the same mesh of subsets in Figure \ref{fig:prompting:hypoonly}. For every subset, the H-only model still trails behind the S\&S classifiers. These gaps are partly explained by the fact that the H-only model does not capture all existing artefacts, but is merely a proxy to their strengths. 

As shown, the \textit{Directional} subset indeed has particularly strong artefacts to exploit, with the H-only model reaching 48.9\% $\it AUC_{norm}$, far above the other subsets. Between \textit{Paraphrases-DirTrue} and \textit{Paraphrases-DirFalse}, the relative performance of S\&S model on them is aligned with their relative strengths of artefacts; this means, for RoBERTa, \textit{Paraphrases} is in fact similarly separable from \textit{DirTrue} and \textit{DirFalse}, as is in line with expectation.

Also interesting is the comparison between
the \textit{directional} and \textit{symmetric} subsets. The two subsets had similar S\&S performances; however, there is a difference of over 20\% between their hypothesis-only artefact strengths. That means the S\&S classifier is after all far better at the \textit{symmetric} subset than the \textit{directional} one.

For a crude ranking, we inspect each subset according to the FullModel-HOnly ratio by $AUC_{norm}$ (lower the stronger artifacts). We find at rock bottom the \textit{Paraphrases-DirFalse} and \textit{DirTrue-DirFalse} subsets with this ratio at 1.55 and 1.70 respectively, indicating that their full-model scores are the heaviest over-estimations; next up is the 2.10 for \textit{DirFalse-Unrelated}, all the other subsets have this ratio around 3.


\subsection{The Story behind Artefacts - A Summary}

From the S\&S triangle in Figure \ref{fig:prompting:primary} and the artefact triangle in Figure \ref{fig:prompting:hypoonly}, we make the findings below. 

We find that \textit{DirTrue} and \textit{DirFalse} look similarly separable from both \textit{Paraphrases} and \textit{Unrelated}. Moreover, for \textit{DirTrue} and \textit{DirFalse}, it is consistently harder to separate them from \textit{Paraphrases} than from \textit{Unrelated}. This suggests RoBERTa is modelling both of them as closer to \textit{Paraphrases} than to \textit{Unrelated}. This is typical for symmetric measures, but also possible for directional ones.

However, as we examine the directionality triangles, we find that RoBERTa shows poor directionality in \textit{Paraphrases}-\textit{DirX} subsets. It also does far worse on the \textit{directional} subset than on
\textit{symmetric}, but the $\it AUC_{norm}$ is inflated by dataset artefacts.

In conclusion, we find evidence that RoBERTa (and BERT, see Appendix \ref{supp:lh_result}) are poor learners of directional inference. However, $\it AUC_{norm}$ values are, after all, comparable but not additive, so we still cannot measure the net competence by deducting H-only scores from S\&S scores. Thus, we cannot conclusively prove or falsify claims about directionality of supervised LM methods. For robust extrinsic evaluation, we present BoOQA dataset.

\section{Boolean Open QA (BoOQA) Dataset}
\label{Sec:booqa}

We have shown that LevyHolt \textit{directional} subset is especially infested with dataset artefacts and is not a robust benchmark for supervised methods. With the directionality triangle, we have shown evidence that LMs have limited ability to learn directional entailments. However, this experiment is tedious and inconclusive: even solving the triangle does not guarantee solving directionality, as artefacts always persist;
besides, we are still unable to compare between supervised and unsupervised methods. 

To establish an extrinsic, robust and realistic evaluation benchmark for directional predicate entailments, and to encourage future research on the topic, we present the Boolean Open-QA dataset (BoOQA).

BoOQA is an automatically constructed dataset in English and Chinese. It draws inspirations from the \textit{machine reading at scale} task \cite{chen_reading_2017} as well as the QA evaluation in \cite{mckenna_multivalent_2021}. The task is formalized as: given a proposition and a large pool of context articles, determine whether the queried proposition is true based on the context pool.\footnote{We consider only propositions with binary relations; unary relations have only one argument for context disambiguation, thus are naturally noisier.}

\subsection{Dataset Elicitation}
\label{Sec:booqa:data_elicitation}

In principle, we follow \citet{mckenna_multivalent_2021} in building this dataset. We separate large news corpora into context windows of 3-day time spans, select the frequent argument pairs as starring arguments, then select frequent relations about these starring arguments as positive propositions. We generate negatives from positives, by keeping the argument pairs unchanged, and selecting negative predicates from the WordNet hyponyms of positive predicates. For instance, for a positive proposition ``\textcolor{blue}{John} \textcolor{red}{goes to} \textcolor{olive}{\sc ikea}'', we may derive a negative proposition ``\textcolor{blue}{John} \textcolor{red}{drives to} \textcolor{olive}{\sc ikea}''.

Negative predicates are hyponyms that are felicitous but wrong in the current context: we check if a hyponym is felicitous by its corpus-wide presence; we check if a hyponym is wrong for the current context by its context-window-wide absence with the corresponding argument pair \footnote{This is based on the assumption that a comprehensive pool of news articles would follow the Gricean cooperative principle of communication \cite{davis_implicature_2019}, that it would include all and only the known facts.}.

WordNet hyponyms are good candidate negative predicates because: 1) they are guaranteed to be similar to their corresponding positives, thus adversarial to symmetric similarity; 2) they are more specific than their positives and less likely to be true, so the risk of getting false negatives is reduced.

We take samples from the positive and negative propositions, and blend them to create the BoOQA dataset. For evaluation, we take these propositions as hypotheses, and retrieve premises from the corresponding context windows\footnote{To avoid producing misleadingly optimistic scores, we force all methods to ignore the exact sentences from which the hypotheses (or their corresponding hypernyms) are extracted.}. 
The task is then to judge the truthfulness of the hypotheses given the context windows. For a good entailment measure, positive hypotheses should be entailed by context more confidently than negatives, where context means other mentions of the same starring arguments within the same context window.

For English, we follow \citet{mckenna_multivalent_2021} in using NewsCrawl \cite{barrault_findings_2019} as our source corpus; for Chinese, we use the CLUE news corpus \cite{xu_clue_2020}. As we include entailment graphs as unsupervised baselines, we remove from these corpora the articles seen in entailment graph induction corpora; on the other hand, we refrain from deleting overlaps with the much larger LM-pretrain corpora, to avoid biasing our dataset toward non-mainstream coverage.

Given the analogy between our BoOQA dataset and the evaluation in \citet{mckenna_multivalent_2021}, we leave most details of dataset construction to Appendix \ref{supp:booqa_nick} and refer readers also to their paper. Here we highlight our additional efforts for the quality and robustness of BoOQA.

\paragraph{Quality}

When generating negative predicates, we look for hyponym replacements for every span in the positive predicates, instead of just the head verbs. 
For example, for ``play game with'', we can generate candidate negative ``\textbf{foul} game with'' as well as ``play \textbf{practice game} with''. The former may be absent from the whole corpus, the latter may be present somewhere in the corpus but not the current context window; then the former would be discarded and the latter kept. 
This allows us to generate negatives for more complex predicates, avoiding biasing toward shorter, simpler predicates; it is especially useful in Chinese, where word boundaries are unmarked and WordNet entries are present across multiple levels of granularity.

When collecting hyponyms of WordNet matches, instead of looking into the first synset of each match, for English we disambiguate matches to their most relevant synsets with a WSD method \cite{yap_adapting_2020}. The WSD method takes in a span in sentential context as input, and produces the most likely synset for that span given the context.
This way, we find more accurate hyponyms and generate more challenging negatives that are similar to their positives. Unfortunately, no such models or datasets are available for Chinese, so we have to fall back to using the first synset.

Thirdly, we raise the bar for candidate negatives to be considered ``wrong in the current context'': not only must the predicate itself be absent from the context window (with the current arguments), but all its WordNet synonyms must also be absent. This stricter filter further reduces false negatives.

Fourthly, we expand our source corpora to entire NewsCrawl / CLUE instead of their slices, so we avoid biasing toward any time spans. With larger corpora, we impose larger thresholds for starring-arguments, so the events in question are thoroughly discussed, fewer positives cannot be inferred from the context, namely fewer false positives\footnote{False positives are still true events, but models should not be expected to label them as ``positive'' given the context.}.

Lastly, we sample data entries in bundles of positives and negatives. Whenever a positive is sampled, its corresponding negatives must also be sampled, and vice versa. This strengthens our dataset in the way that each method needs to assign higher confidence values to positives than their corresponding negatives to get good results.

\paragraph{Robustness}
While inspecting the QA dataset in \citet{mckenna_multivalent_2021}, we find that their positive and negative propositions are separable even without context: the generated negatives simply look less plausible than their corresponding positives. 
We again use hypothesis-only (H-only) models to quantify this effect. Since we use the context to retrieve premises, the H-only model by definition means ``without context''. Similarly to \citet{schmitt_language_2021}, we sample train / dev2 sub-splits out of the dev set for fine-tuning.\footnote{For details of this H-only model see Appendix \ref{supp:qaeval_hypoonly}.} 

Ideally, without context, the negatives should be inseparable from the positives, the H-only model should do badly. However, on McKenna dataset, it got 78.3\% $\it AUC_{norm}$. That means McKenna dataset is also infiltrated by spurious correlations. Although this is not fatal since the dataset is only for evaluation so models are not exposed to such artefacts, it is still desirable to improve robustness by bringing the negatives closer to the positives.

To this end, we also raise the bar for candidate negatives to be considered ``felicitous'': they have to be not just present in the corpus, but also frequent enough. We set the same frequency thresholds for felicitousness between positives and negatives.

In addition, when sampling from the populations, we create negatives samples such that they have similar frequency distributions as the positives population. We calculate the ratio of predicates in positives population falling into each frequency range, then sample the negatives proportionally.
\footnote{We sample positives/negatives in bundles, so the positive samples are generally more frequent than the population.} 

\begin{table}[t]
    \centering
    \begin{tabular}{|c|c|c|}
    \hline
    Train / Dev Set & Test Set & $\it AUC_{norm}$ \\\hline
    $\it BoOQA_{EN}$ & $\it BoOQA_{EN}$ & 51.0 \\
    $\it BoOQA_{ZH}$ & $\it BoOQA_{ZH}$ & 84.2 \\
    \it{McKenna} & \it{McKenna} & 78.3 \\\hline
    $LevyHolt_{Full}$ EN & $\it BoOQA_{EN}$ & 2.8 \\
    $LevyHolt_{Dir}$ EN & $\it BoOQA_{EN}$ & 1.1 \\
    $LevyHolt_{Sym}$ EN & $\it BoOQA_{EN}$ & 0.8 \\\hline
    $LevyHolt_{Full}$ ZH & $\it BoOQA_{ZH}$ & 4.0 \\
    $LevyHolt_{Dir}$ ZH & $\it BoOQA_{ZH}$ & 1.2 \\
    $LevyHolt_{Sym}$ ZH & $\it BoOQA_{ZH}$ & 2.4 \\\hline
    \end{tabular}
    \caption{Robustness measured by H-only model performance, values in \%, lower is better. McKenna is the QA dataset in \citet{mckenna_multivalent_2021}, $LevyHolt_{Dir}$ and $LevyHolt_{Sym}$ are \textit{directional} and \textit{symmetric} subsets.}
    \label{Tab:qaeval_hypo}
\end{table}

We apply H-only model to BoOQA and report the results in Table \ref{Tab:qaeval_hypo}. $\it BoOQA_{EN}$ enjoys a reasonably significant drop against McKenna dataset for H-only performance, meaning less exploitable spurious correlations. However, we acknowledge that generated negatives are naturally hard to blend with collected positives, and there remain clues in the hypothesis themselves indicating their truthfulness. For $\it BoOQA_{ZH}$, the H-only result is still high even with our above efforts. Fortunately, for both languages, H-only models trained on LevyHolt subsets cannot identify the truthfulness of BoOQA hypotheses, indicating that BoOQA remains safe as an extrinsic evaluation benchmark.

\begin{table}[h]
    \centering
    \begin{tabular}{|c|c|c|}
    \hline
        \# of & positives & negatives \\\hline
        $BoOQA_{EN}$ dev & 21809 & 36782 \\
        $BoOQA_{EN}$ test & 21773 & 36755 \\\hline
        $BoOQA_{ZH}$ dev & 18855 & 31117 \\
        $BoOQA_{ZH}$ test & 19092 & 31479 \\\hline
    \end{tabular}
    \caption{Number of positive / negative entries in each subset of BoOQA.}
    \label{Tab:booqa_stats}
\end{table}

\paragraph{Statistics and Discussions}

We provide dev/test splits in BoOQA, each has circa 20k positives and 40k negatives, as shown in Table \ref{Tab:booqa_stats}. Positives and negatives are sampled in bundles where at most 2 negatives correspond to each positive (subject to availability). 

Notably, the size of 60K entries is chosen empirically, as we find results to have already stabalized at this size. Though the number of entries is 60K, the number of entailments to be checked is magnitudes larger: in assigning a score to each entry, a method must inspect the relation from each piece of evidence to the proposition at question. Moreover, since the construction process is fully automatic, it is easily scalable and extendable to other corpora.

In Appendix \ref{supp:booqa_case_study}, we also present a human analysis for the felicitousness of BoOQA entries, and show that 85\%+ correctness can be expected for both labels and both languages, which is relatively high and even comparable to some crowd-annotated datasets.

On BoOQA, directional entailments are not needed to perform above random, but are needed to reach high precisions. Consider a positive proposition $\textcolor{red}{A}$ and its corresponding negative $\textcolor{red}{B}$; a third proposition $\textcolor{red}{\it C}$ is retrieved from context articles. In a simplified sense, a model is correct as long as it scores $\textcolor{red}{\it C} \vDash \textcolor{red}{A}$ higher than $\textcolor{red}{\it C} \vDash \textcolor{red}{B}$.
Now suppose there are two perfect measures of entailments, except that one is directional, the other symmetric. When neither queried propositions entail the context (i.e. $\textcolor{red}{A} \nvDash \textcolor{red}{\it C}$ and $\textcolor{red}{B} \nvDash \textcolor{red}{\it C}$), the two measures behave similarly; otherwise, when some of these reverse-direction entailments hold, chances are that the symmetric measure would fall into the trap and produce spurious results. Through this property, 
directionality is involved in BoOQA dataset; we offer a detailed analysis in Appendix \ref{supp:booqa_directionality}.

Moreover, although BoOQA negatives are built off WordNet hyponyms of positives, the directional entailments tested are not confined to hyponymy/hypernymy. This is because BoOQA tests for entailment strengths from context-propositions to the queries, instead of between the positive (hypernym) and the negative (hyponym) queries themselves. This allows us to attend to all kinds of directional entailments as is present in the general news domain, for instance, precondition-consequences.


\subsection{Experiments and Baselines}
\label{Sec:booqa:baselines}

In Table \ref{Tab:qaeval}, we present baselines on BoOQA test set with 3 types of methods: unsupervised LMs, supervised LM-driven methods and entailment graphs. 

Each method retrieves supporting evidence from context, which are the extracted propositions in the same context window involving the same arguments as the queried hypothesis; they then calculate a score from each piece of evidence to the hypothesis, and take the highest score w.r.t any evidence as the final confidence score for a hypothesis.

\noindent \textbf{Unsupervised LMs} They compute cosine similarities between BERT-base encodings of the retrieved supporting evidence and the hypothesis, as confidence scores. We also tried BERT-large, RoBERTa as well as alternative retrieval approaches, we leave them to Appendix \ref{supp:booqa_experiments}, as their dev set results are not superior to the condition reported.

\noindent \textbf{Supervised LMs} They use $S\&S_{subset}$ models with RoBERTa-base to compute entailment scores from retrieved evidence to the hypotheses. 
We present $S\&S_{full}$ as supervised LM baseline for $\it BoOQA_{EN}$ and $\it BoOQA_{ZH}$. On $\it BoOQA_{EN}$, we also compare $S\&S_{Directional}$, $S\&S_{Symmetric}$ and $\widehat{S\&S}_{Full}$ (trained on full LevyHolt with symmetric prompts, see \S\ref{Sec:prompting:ss_triangle}). On $\it BoOQA_{EN}$ we also provide large model results in Appendix \ref{supp:booqa_experiments}; those are not higher than with base model. 

\noindent \textbf{Entailment Graphs} They are 
unsupervised entailment detection methods, where nodes in each graph are predicates, 
and directed edges between nodes represent entailment relations between predicates. Entailment graphs are induced from textual corpora with directional co-occurrence signals. At test time, entailment scores are collected by looking up the edge weights from each predicate-in-evidence to the predicate-in-hypothesis. For English, we evaluate with BInc \cite{hosseini_learning_2018}, CNCE \cite{hosseini_open-domain_2021} and EGT2 \cite{chen_entailment_2022}; for Chinese, we evaluate with BInc \cite{li_cross-lingual_2022}.


\begin{table}[t]
    \centering
    \begin{tabular}{|c|c|c|}
        \hline
        $\it AUC_{norm}$ (\%) & $BoOQA_{En}$ & $BoOQA_{Zh}$  \\\hline
        $\it LM_{unsupervised}$ & $15.9$ & $30.6$ \\
        $S\&S_{Full}$ & $25.6$ & $23.1^{*}$ \\
        $\widehat{S\&S}_{Full}$ & $26.2$ & - \\
        $S\&S_{Symmetric}$ & $25.1$ & - \\
        $S\&S_{Directional}$ & $15.1$ & - \\\hline
        EG BInc & $29.8$ & $39.5$ \\
        EG CNCE & $\textbf{34.5}$ & - \\
        EG EGT2 & $26.8$ & - \\\hline
    \end{tabular}
    \caption{Baselines on BoOQA test set in English and Chinese.
    All methods are taken out-of-the-box. ``EG XX'' are entailment graphs with various entailment scores as described below.}
    \label{Tab:qaeval}
\end{table}

On $BoOQA_{EN}$, we re-examine the $S\&S$ gauge models in comparison to various unsupervised methods.
$\it LM_{unsupervised}$ does poorly, yet $S\&S_{Directional}$ does even worse.
This verifies that its judgements are polluted by dataset artefacts. Moreover, it does not matter whether $S\&S$ is exposed to only $symmetric$ entries or uses purely symmetric prompts, the performance is almost identical. This means RoBERTa is not learning from directional supervision (in LevyHolt) the directional ability to make progress on BoOQA.
These results echo our findings in \S\ref{Sec:prompting}.
Though being supervised and only with domain transfer, the $S\&S$ models fail to outperform fully unsupervised entailment graphs on BoOQA.


\begin{figure}[t]
    \centering
    \includegraphics[scale=0.45]{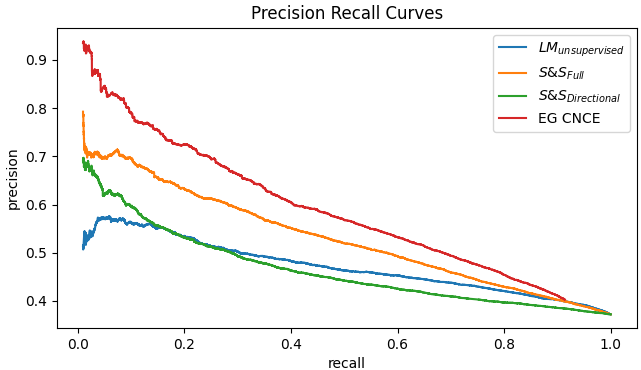}
    \caption{Precision-recall curves for the representative methods of each class on $\it BoOQA_{EN}$ test set.}
    \label{Fig:booqa_pr_rec_curves}
\end{figure}

Entailment graph methods generally have decent performance. EG CNCE enjoys a considerable advantage, particularly at the higher-precision end. While it may seem surprising for an unsupervised method to prevail in high-precisions range, this is expected given that in BoOQA, directionality is required for methods to reach high precisions. On the other hand, these discrete entailment graphs are hampered by sparsity; although a score is retrieved for most queries with the help of fuzzy matching\footnote{Ignoring semantic role labels for arguments in the entailment graph representations of predicates; see Appendix \ref{supp:booqa_experiments} for more details and a comparison of Pr-Rec curves.}, the high precisions are short-lived measured by recall, dropping rapidly to the moderate range of sub-60\%. This suggests generalising unsupervised directional entailment graphs to continuous encoding spaces could be a promising direction to follow.


\section{Related Work}
\label{Sec:related_work}

Related work to our paper broadly fall into three categories: Language-Model prompting, entailment graphs and natural language inference.

LM-prompting methods convert input in target tasks into natural language sentences (prompts), thereby converting target tasks into either language modelling or sentence classification. \citet{petroni_language_2019} probed LM for factual knowledge, where \citet{jiang_how_2020} proposed automatic prompt discovery; on the other hand, GPT-3 \cite{NEURIPS2020_1457c0d6} showed large LMs can do few-shot / zero-shot transfer with prompts, and \citet{schick_its_2021} showed a similar effect with smaller LMs and natural language prompts. \citet{liu_pre-train_2021} provided a comprehensive review of LM-prompting, which we refer readers to for more details. 

Entailment graphs (EGs) are graphs of predicates where directed edges represent entailment relations between pairs of predicates. Weights on edges are entailment scores, induced from large textual corpora via unsupervised learning, based on directional co-occurrence signals; examples include BInc \cite{szpektor_learning_2008,hosseini_learning_2018}, link prediction \cite{hosseini_duality_2019} and the LM-driven CNCE \cite{hosseini_open-domain_2021}. Being fully unsupervised, EGs are sheltered from artefacts in training data, but are at the same time subject to noisy co-occurrence signals and sparsity common to discrete language resources.

Notably, while CNCE EG \shortcite{hosseini_open-domain_2021} uses LM, it involves contextualized encodings for premises, and uncontextualized encodings for hypotheses, trained from scratch with millions of unsupervised data entries. The architecture and abundant supervision enables CNCE to encode directional entailments. EGT2 \cite{chen_entailment_2022} is another LM-driven EG contemporary to our paper, with intermediate fine-tuning on MNLI \cite{williams_broad-coverage_2018} and PPDB \cite{ganitkevitch_ppdb_2013} annotations. They report good resuts on the \textit{directional} subset of LevyHolt\footnote{Their reported directional subset is different from the original one in \citet{holt_probabilistic_2019}, see Appendix \ref{supp:egt2} for details.}, but perform mediocrely on BoOQA. We speculate that this could be due to instabilities of the small LevyHolt \textit{directional} test set.


Natural Language Inference is concerned with the general task of ``Does sentence X entail sentence Y?''. Classic NLI datasets such as SNLI \cite{bowman_large_2015} or MNLI \cite{williams_broad-coverage_2018} are populated with entries involving hypernyms of nouns, logic reasoning like $A \wedge B \rightarrow B$ and ``$A$ under the condition of $B$'' $\rightarrow A$, whereas predicate entailments are rarely involved. Predicate entailment datasets like LevyHolt \cite{levy_annotating_2016,holt_probabilistic_2019}, SherlIic \cite{schmitt_sherliic_2019} or our BoOQA dataset, are a sub-class of NLI datasets focused on inference with predicates, where the arguments and adjuncts are carried over from premise to hypothesis. 
In particular, LevyHolt and our BoOQA dataset both feature the property of directionality, where the entailment is valid in one direction, but invalid in the other.

General NLI datasets are prune to simple dataset artefacts such as length, lexical overlaps or hypothesis-only models \cite{gururangan_annotation_2018,poliak_hypothesis_2018,kalouli_annotation_2021,mccoy_right_2019}. These echo our observations on predicate entailments, and underscore the importance of a robust and extrinsic evaluation.

\section{Conclusion}
In this paper, we show that existing LM prompting methods show limited ability to learn directional entailments, and instead overfit to dataset artefacts in the LevyHolt \textit{directional} subset. We present BoOQA, a robust and extrinsic multilingual evaluation benchmark on directional predicate entailments as a Boolean Open QA dataset. On BoOQA, we show that LM prompting methods do not learn to pay attention to directionality from directional supervision, while the entailment graph methods remain limited by sparsity. We point to the generalisation of entailment graphs to continuous spaces and directionality-oriented intermediate pre-training of LMs as directions of future work.

\section*{Limitations}
The remaining limitations of our work are two-fold. Firstly, our experiments are done primarily with RoBERTa and BERT models. This choice is taken following the positive results reported in \citet{schmitt_language_2021,schmitt_continuous_2021}. We argue that the consistency among these models indicates a broader generalisability of our conclusions, however we are unfortunately unable to enumerate over the wide variety of language models to prove this point. Particularly, due to limits to our computational resources, we are unable to experiment with extremely large language models such as GPT-3. We release the code and data for the experiments as described in \S\ref{Sec:prompting} and \S\ref{Sec:booqa}, and encourage the community to try out any language models they might be interested in.

The second aspect of limitation is, despite our efforts to improve the robustness of our BoOQA dataset, there remain correlations between just the surface forms of the propositions and their truthfulness, which could be exploitable under a supervised setting. We argue that one cannot eliminate such correlations without heavily manipulating which propositions are of interest to the dataset, which is against our aim of preserving the natural distribution of propositions in large-scale news corpora; we also argue that this is harmless as long as BoOQA remains an extrinsic benchmark, namely, the training set of any supervised method should not have the same kinds of correlations. We have checked that the LevyHolt dataset and its various subsets satisfy this criterion; we advise that readers attempting this dataset with their own methods do the same check to ensure comparable results.

\section*{Acknowledgements}
The authors would like to thank Jeff Pan and Nick Mckenna for helpful discussions and the anonymous reviewers for their valuable feedback. This work was supported partly by ERC Advanced Fellowship GA 742137 SEMANTAX, a Mozilla PhD scholarship at Informatics Graduate School and the University of Edinburgh Huawei Laboratory.

\bibliography{main.bib}
\bibliographystyle{acl_natbib}

\appendix

\section{Graphic Illustration for $\it AUC_{norm}$}
\label{supp:auc_norm_demo}

\begin{figure}[h!]
    \centering
    \includegraphics[scale=0.4]{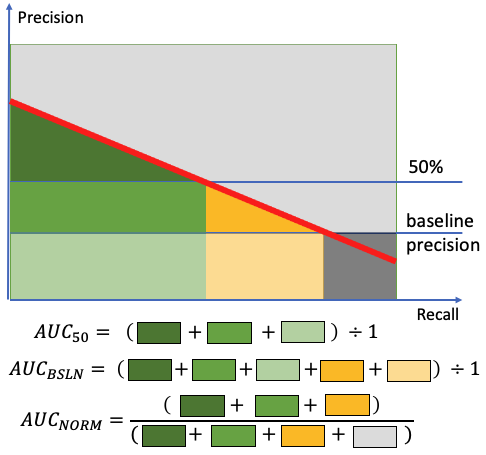}
    \caption{Diagram illustration of $\it AUC_{norm}$ in comparison to other $\it AUC$ metrics. Red and bolded is a simplified precision-recall curve, each $\it AUC$ metric is represented as ratios of area of the respective coloured blocks. The total area of all blocks equals to 1. Note that in practice the precision of any method rarely comes below random (as that would mean the least confident entries carry some kind of reverse judgements), and in any case not beyond the scale reasonable for random fluctuations. That is, real pr-rec curves do not cut deeply into the dark grey area like the over-simplified demo.}
    \label{Fig:auc_norm_demo}
\end{figure}

See Figure \ref{Fig:auc_norm_demo} for a graphic illustration of $\it AUC_{norm}$, and a comparison between different metrics.

\section{Details and Comparison of S\&S Models}
\label{supp:ss_probe}

\subsection{Between Discrete / Continuous Prompts}

The S\&S models \cite{schmitt_language_2021, schmitt_continuous_2021}  are LM-prompting models based on RoBERTa. Of which, S\&S-discrete \cite{schmitt_language_2021} falls into the category of ``Fixed-prompt LM Tuning'' according to the taxonomy in \citet{liu_pre-train_2021}; whereas its follow-up work S\&S-continuous \cite{schmitt_continuous_2021} falls into the category of ``Prompt+LM Tuning''.

S\&S-discrete uses human-engineered natural language prompts. Given a premise-hypothesis pair, it instantiates each prompt with the pair, feeds them into the LM, thus converts the task into a regular binary classification task from the LM representations of the instantiated prompts. 

Notably, we discovered a bug in the implementation of \citet{schmitt_language_2021}: due to missing negations, for experiments on LevyHolt dataset, the manual prompt (d) and (e) listed in \citet{schmitt_language_2021} were effectively converses of prompt (c) and (a) respectively. After fixing the bug, we found the model performance with RoBERTa-base, measured by $\it AUC_{50\%}$, as reported in their original paper, rises from 76.9\% to 80.7\%; correspondingly, the $\it AUC_{norm}$ value rises from 75.3\% to 77.7\%.

\begin{table}[ht]
    \centering
    \begin{tabular}{|c|c|c|c|}
    \hline
        Model & LevyHolt & SherlIic \\\hline
        \multicolumn{3}{|l|}{Trained with LevyHolt} \\\hline
        S\&S-discrete & 80.7 (77.7) & 69.5 (59.5) \\
        S\&S-continuous & 79.2 (76.2) & 65.1 (54.3) \\\hline
        \multicolumn{3}{|l|}{Trained with SherlIic} \\\hline
        S\&S-discrete & 35.5 (47.7) & 63.8 (51.5) \\
        S\&S-continuous & 25.1 (32.8) & 66.5 (57.9) \\\hline
    \end{tabular}
    \caption{Generalization experiment between LevyHolt and SherLIic datasets, examined models are S\&S-discrete and S\&S-continuous. Performance is measured in $\it AUC_{50\%}$, where the $\it AUC_{norm}$ values are appended in brackets. Note that reported values are from our reproduction with our 100 sets of hyper-parameter configs, so the exact values slightly differ from the originals.}
    \label{tab:levy_sherliic_generalization}
\end{table}

S\&S-continuous is in essence similar to S\&S-discrete, except that it replaces the embeddings of human-engineered natural language prompt tokens with sequences of trainable continuous embeddings. S\&S-continuous reported SOTA results, outperforming S\&S-discrete; however, we stick with the S\&S-discrete model for our experiments for the following reasons:

\begin{itemize}[itemsep=0pt]
    \item After fixing the bug in S\&S-discrete implementation, its genuine performance is not worse than that reported for S\&S-continuous;
    \item We did a generalization experiment between the two predicate entailment datasets, LevyHolt \cite{levy_annotating_2016} and SherlIic \cite{schmitt_sherliic_2019}, where we found that S\&S-discrete is better at generalizing from one dataset to the other, as shown in Table \ref{tab:levy_sherliic_generalization}; that suggests the S\&S-discrete is less prone to overfitting;
    \item For the same LM backbone, S\&S-discrete has a smaller computational footprint.
\end{itemize}

\subsection{Subsets with Same-label Sub-groups}

\begin{table}[ht]
    \centering
    \begin{tabular}{|c|cc|}
        \hline
        $\it AUC_{norm}$ (\%) & Discrete & Continuous \\\hline
        Para-DirTrue & 26.9 & 25.5 \\\
        DirFalse-Unrl & 58.6 & 62.2 \\\hline
    \end{tabular}
    \caption{Performance of $S\&S$ model on \textit{Paraphrases-DirTrue} and \textit{DirFalse-Unrelated} subsets. \textit{Discrete} and \textit{Continuous} are S\&S-discrete and S\&S-continuous.}
    \label{Tab:dirsym_sscont}
\end{table}

For the subsets with same-label sub-groups, namely, the subsets whose two sub-groups are either both true entailments or both non-entailments, we assign positive labels to the sub-group with higher level of relevance, and negative labels to the other. As discussed in \S \ref{Sec:prompting}, we acknowledge that \citet{schmitt_language_2021} uses hand-crafted prompts which are designed for labelling the truthfulness of a candidate entailment, so they may not be ideal for separating same-label sub-groups. While we argue that fixed-prompt LM tuning models are not too sensitive to their specific prompts \cite{logan_iv_cutting_2021,webson_prompt-based_2022}, we also supplement the experiment with continuous prompt models.

\begin{table}[t]
    \centering
    \begin{tabular}{|c|c|c|c|}
        \hline
        \diagbox[width=2.3cm, height=1.4cm]{Train/Dev}{Test} & Directional & Symmetric & Full \\\hline
        Directional & 71.1 & 2.6 & 11.6 \\\hline
        Symmetric & 2.1 & 90.1 & 70.9 \\\hline
        Full & 52.4 & 90.7 & 83.7 \\\hline
    \end{tabular}
    \caption{Generalization performance of \textbf{RoBERTa-large} S\&S classifier on the \textit{DirTrue-DirFalse} (Directional) and \textit{Paraphrases-Unrelated} (Symmetric) subsets of LevyHolt. Values are in \% of $\it AUC_{norm}$. Rows are train/dev subsets, columns are test subsets.}
    \label{Tab:dir_and_sym_roberta_large}
\end{table}

\begin{table}[t]
    \centering
    \begin{tabular}{|c|c|c|}
    \hline
    \rule{0pt}{15pt} $\it AUC_{norm}$ & $S\&S$ & $\widehat{S\&S}$ \\\hline
    Para-DirTrue & 43.4 & 33.2 (-10.2\%) \\
    Para-DirFalse & 59.4 & 49.9 (-9.5\%) \\\hline
    \end{tabular}
    \caption{Comparison between \textbf{RoBERTa-large} $S\&S$ classifiers (regular) and $\widehat{S\&S}$ (symmetric prompts). \textit{Paraphrases-DirTrue} and \textit{Paraphrases-DirFalse} subsets are concerned.}
    \label{Tab:symp_roberta_large}
\end{table}

The prompts in S\&S-continuous are sequences of trainable embeddings, which can be tuned for any purpose, not just to separate true and false entailments; moreover, as Table \ref{tab:levy_sherliic_generalization} shows, S\&S-continuous and S\&S-discrete have similar performance when tested on the same dataset as train set. So we use S\&S-continuous as a control model, to understand how much the S\&S-discrete is affected by prompt suitability. 

In Table \ref{Tab:dirsym_sscont}, we report results on the two same-label subsets with S\&S-continuous in comparison to S\&S-discrete. We find the results from the two models to be very similar. This means, as aligned with our argument above, the prompt-suitability issue is not jeopardising our experiment.


\section{LevyHolt Experiments on More LMs}
\label{supp:lh_result}

In this section, we report more results from the LevyHolt experiments. In particular, we replicate the generalisation experiment and the symmetric-prompt experiment in \S\ref{Sec:prompting:ss_triangle} with RoBERTa-large, BERT-base and BERT-large models.

\begin{table}[t]
    \centering
    \begin{tabular}{|c|c|c|c|}
        \hline
        \diagbox[width=2.3cm, height=1.4cm]{Train/Dev}{Test} & Directional & Symmetric & Full \\\hline
        Directional & 75.7 & 4.4 & 15.4 \\\hline
        Symmetric & 0.3 & 67.5 & 53.0 \\\hline
        Full & 30.1 & 69.5 & 62.4 \\\hline
    \end{tabular}
    \caption{Generalization performance of \textbf{BERT-base} S\&S classifier on the \textit{DirTrue-DirFalse} (Directional) and \textit{Paraphrases-Unrelated} (Symmetric) subsets of LevyHolt. Values are in \% of $\it AUC_{norm}$. Rows are train/dev subsets, columns are test subsets.}
    \label{Tab:dir_and_sym_bert_base}
\end{table}

\begin{table}[t]
    \centering
    \begin{tabular}{|c|c|c|}
    \hline
    \rule{0pt}{15pt} $\it AUC_{norm}$ & $S\&S$ & $\widehat{S\&S}$ \\\hline
    Para-DirTrue & 5.0 & 3.4 (-1.6\%) \\
    Para-DirFalse & 17.8 & 11.7 (-6.1\%) \\\hline
    \end{tabular}
    \caption{Comparison between \textbf{BERT-base} $S\&S$ classifiers (regular) and $\widehat{S\&S}$ (symmetric prompts). \textit{Paraphrases-DirTrue} and \textit{Paraphrases-DirFalse} subsets are concerned.}
    \label{Tab:symp_bert_base}
\end{table}

\begin{table}[t]
    \centering
    \begin{tabular}{|c|c|c|c|}
        \hline
        \diagbox[width=2.3cm, height=1.4cm]{Train/Dev}{Test} & Directional & Symmetric & Full \\\hline
        Directional & 71.7 & 2.7 & 9.1 \\\hline
        Symmetric & 0.0 & 77.1 & 58.2 \\\hline
        Full & 49.4 & 80.9 & 75.4 \\\hline
    \end{tabular}
    \caption{Generalization performance of \textbf{BERT-large} S\&S classifier on the \textit{DirTrue-DirFalse} (Directional) and \textit{Paraphrases-Unrelated} (Symmetric) subsets of LevyHolt. Values are in \% of $\it AUC_{norm}$. Rows are train/dev subsets, columns are test subsets.}
    \label{Tab:dir_and_sym_bert_large}
\end{table}

\begin{table}[t]
    \centering
    \begin{tabular}{|c|c|c|}
    \hline
    \rule{0pt}{15pt} $\it AUC_{norm}$ & $S\&S$ & $\widehat{S\&S}$ \\\hline
    Para-DirTrue & 10.8 & 20.2 (+9.4\%) \\
    Para-DirFalse & 54.9 & 37.4 (-17.5\%) \\\hline
    \end{tabular}
    \caption{Comparison between \textbf{BERT-large} $S\&S$ classifiers (regular) and $\widehat{S\&S}$ (symmetric prompts). \textit{Paraphrases-DirTrue} and \textit{Paraphrases-DirFalse} subsets are concerned.}
    \label{Tab:symp_bert_large}
\end{table}

\textit{Directional-Symmetric} generalization results for the three LMs are summarized in Table \ref{Tab:dir_and_sym_roberta_large}, \ref{Tab:dir_and_sym_bert_base} and \ref{Tab:dir_and_sym_bert_large} respectively, mirroring Table \ref{Tab:dir_and_sym}. The conclusions are consistent, where S\&S classifiers trained on the two subsets are still not generalising to each other.

The comparison with symmetric-prompt baselines for the above three LMs are in Table \ref{Tab:symp_roberta_large}, \ref{Tab:symp_bert_base} and \ref{Tab:symp_bert_large}. There again, scaling up the language model or switching from RoBERTa to BERT cannot lift the regular-prompt S\&S model further away from the symmetric-prompt one. Particularly in Table \ref{Tab:symp_bert_large}, $\widehat{S\&S}_{Para-DirTrue}$ actually outperforms $S\&S_{Para-DirTrue}$, highlighting that performances of LMs are unstable on smaller fine-tune datasets, even with 100 sets of hyper-parameters for choice.

These observations echo those listed in \S \ref{Sec:prompting}, and are verify that our conclusions in \S \ref{Sec:prompting} are generalisable at least to RoBERTa-large and BERT language models. We further argue that they are very likely generalisable to other language models following the same training scheme as well.

\section{Detailed Guide for BoOQA Construction}
\label{supp:booqa_nick}

This section is of interest to readers who wish to dig into the details of how BoOQA is built, or to build their own flavour of BoOQA dataset.

The BoOQA dataset is constructed off large-scale general-domain news corpora. For English, we use the NewsCrawl corpus, which contains 7.8M news articles published between 2007 and 2017; for Chinese, we use the CLUE corpus, which contains 2.4M news articles, all published in 2016. With corpora at such scale, they naturally have overlaps with the corpora used for building models. Our introduced baselines include LM-driven methods and Entailment-Graph-based methods. The entailment graphs are induced from smaller corpora, which we can afford to exclude from NewsCrawl or CLUE; on the other hand, the LMs are pre-trained on corpora of much larger scale, where it is impractical to remove all overlaps between the corpora, so we refrain from such attempts. 


We partition each corpus into context windows by disjoint 3-day time spans. We extract open relation triples (subject, predicate, object) from the articles with parsers \cite{reddy_large-scale_2014, li_cross-lingual_2022}. The use of automatic parsing-based relation extraction enables us to collect data entries at large scales. The reason for using these parsers instead of neural taggers such as OpenIE \cite{kolluru_openie6_2020}, is that they have wider coverage and are capable of handling discontinuous predicates, whereas OpenIE only extracts contiguous spans.

From each context window, we look for frequent argument-pairs mentioned in at least 15 different articles with at least 15 different predicates. These argument-pairs star in thoroughly-discussed events in this context window, so that valid propositions about these argument-pairs can be expected to be inferable from existing mentions.

On the other hand, in order to battle the noise introduced by the automatic parsers, and include only the propositions with felicitous predicates in our dataset, we introduce thresholds for predicates to be taken into consideration. Each predicate has to be present with at least 30 different argument pairs for it to be considered felicitous; these can be anywhere in the corpus.

We select propositions between felicitous predicates and starring argument-pairs as the population of positives.\footnote{Population in the sense of statistical population.} Then, we generate adversarial negative propositions from positive propositions.

Negative propositions need to be related to their corresponding positive propositions, but they also need to be false for the current context (i.e. with the current arguments, within the current context window). In practice, we consider a proposition to be false for the current context when it (and all its synonyms) is absent from the articles in the current context window with the current arguments. This is based on the assumption that a comprehensive pool of news articles would follow the Gricean cooperative principle of communication \cite{davis_implicature_2019}, that it would include all and only the known facts.

We generate negatives from WordNet hyponyms: hyponyms are naturally related to their corresponding hypernyms; moreover, they are more specific, so they are less likely to be true. The effect of this property is, when a candidate negative proposition generated from WordNet hyponyms is absent from the current context, it has an even smaller chance of actually being correct for the context. Namely, we are in smaller danger of false negatives.

When selecting negatives from WordNet hyponyms, we iterate over all spans-of-tokens in the seed positive proposition. We check if each span matches any entries in the WordNet; when a match is found, we scan through all its hyponyms, and replace the original span in the proposition with each hyponym, in order to create candidate negatives. For instance, for a positive proposition ``\textcolor{blue}{A} \textcolor{red}{play game with} \textcolor{olive}{B}'', we can replace the span ``play'' with its hyponym ``foul'' to create a candidate ``\textcolor{blue}{A} \textcolor{red}{\textbf{foul} game with} \textcolor{olive}{B}''; we can also replace the span ``game'' with its hyponym ``practice game'' to create a candidate ``\textcolor{blue}{A} \textcolor{red}{play \textbf{practice game} with} \textcolor{olive}{B}''. 

Each candidate negative has to satisfy two criteria to be really considered a negative: it also has to be felicitous, and it has to be absent from the current context. For felicitousness, we set the same threshold as for positives, that is, mentioned with at least 30 different argument pairs anywhere in the corpus. For absence from the current context, we check that not only the predicate in the candidate proposition itself is absent from the current context window, but all its WordNet synonyms must be absent from the current context window as well.

Additionally, for $BoOQA_{EN}$, where we are able to make use of word-sense disambiguation tools \cite{yap_adapting_2020}, for each WordNet match, we select the synset that best suits the context; for $BoOQA_{ZH}$, we select the first synset, since it is the most common synset and has the best chance of being suitable for any given context.

With the positives selected from felicitous predicates between starring arguments, and the negatives generated from the WordNet hyponyms of positives, we have two populations of propositions. We finally create balanced samples from these populations as the resulting BoOQA dataset. We sample no more than two negatives for each positive; we dictate that positives and negatives must be sampled in bundles: no positives can be sampled without their corresponding negatives, and vice versa; we also sample from each context window proportional to its number of articles.

Further, we bring the frequency distributions of positives and negatives closer to each other with bucket sampling. We collect the ratio of predicates in the population of positive propositions falling into each of a pre-defined set of frequency buckets\footnote{The set of bucket frequency boundaries is: (60, 100, 300, 500, 700, 1000, 1500, 2000, 2500, 3000, 4000, 5000, 6000, 8000, 10000, 15000, 20000, 30000, 50000, 100000)}, and sample the negatives in each frequency bucket according to a quota proportional to the ratio of the positive population this bucket represents.


\section{Analysis and Discussions for BoOQA}
\label{supp:booqa_elicitation}

\subsection{Details of H-only Model in BoOQA Robustness Test}
\label{supp:qaeval_hypoonly}

For testing the robustness of our BoOQA dataset, we re-use the H-only model in \S\ref{Sec:prompting:artefacts}, with {(XLM-)}RoBERTa-base LMs. Again we use the word ``true'' or its Chinese equivalent ``正确'' to mask the premise slot in the prompts.

Unlike the LevyHolt H-only setting, in BoOQA, the negative hypotheses are not only not entailed by (any) premise, but also false statements in ground truth; on the other hand, the hypotheses in negative entries of LevyHolt can have any truthfulness in reality. For this reason, if we directly use the hypothesis propositions as training data, the model would be learning not just the artefacts, but also identifying the truthfulness of each statement using its memory from pre-training. To get more precise readings of artefact strengths, we mask the arguments in the propositions with their entity types in FIGER \cite{ling_fine-grained_2012} ontology. This way, we balance between not letting the model know the truthfulness of each proposition and letting the model understand the context of each proposition.

BoOQA comes with only dev / test set. In order to get a train set, we follow \citet{schmitt_language_2021} in sub-splitting the dev set into train / dev2 sets; further, we enforce that these train / dev2 sets are of similar sizes to those in LevyHolt. 

In BoOQA dataset, dev / test sets are sampled from different time spans, and only used for testing. So it is harmless that a hypothesis appears in both dev and test sets, because their truthfulness could be different depending on their different contexts (premises). However, for H-only model, since we no longer use the premises, we do need to eliminate the overlap between train / dev / test splits. We check for overlaps of hypothesis propositions between the data splits and assign them randomly to one set, and remove them from the other. 
Since the same hypothesis can bear different truthfulness values in different context windows, we consider two propositions as overlaps as long as the propositions themselves are identical, regardless of their assigned truthfulness value.

\begin{table}[t]
    \centering
    \begin{tabular}{c}
    \hline
         Felicitous Propositions \\\hline
         Pos: \textcolor{blue}{Mark Zuckerburg} \textcolor{red}{says in} \textcolor{olive}{Facebook} \\
         Neg: \textcolor{blue}{Mark Zuckerburg} \textcolor{red}{responds in} \textcolor{olive}{Facebook} \\\hline
         Infelicitous Propositions \\\hline
         Pos: \textcolor{red}{shown at} \textcolor{blue}{time} \textcolor{red}{for} \textcolor{olive}{time} \\
         Neg: \textcolor{blue}{Police} \textcolor{red}{recites} \textcolor{olive}{suspect} \\\hline
    \end{tabular}
    \caption{Examples of felicitous / infelicitous propositions; ``pos'' denotes positives, ``neg'' negatives.}
    \label{Tab:felicitous_examples}
\end{table}

\subsection{Case Study}
\label{supp:booqa_case_study}

We conduct a case study of dataset quality. Since we expect the premises of the entailment relations to come from large pools of news articles, there are no straight-forward ways of evaluating the label correctness for each context; however, it is possible to evaluate the felicitousness of the hypotheses.

Similarly to above, we again mask the actual arguments with their FIGER \shortcite{ling_fine-grained_2012} types, so human judgements for felicitousness don't get polluted by prior expectations of correctness.

We draw a sample of 100 positive and 100 negatives from both $BoOQA_{EN}$ and $BoOQA_{ZH}$, and manually label their felicitousness. We find the ratio of felicitous samples to be consistent: for English, 86/100 of positive hypotheses are indeed felicitous, 87/100 of negative hypotheses are indeed felicitous; for Chinese, the numbers are 87 for positives and 85 for negatives. 

We list some examples of felicitous and infelicitous entries in Table \ref{Tab:felicitous_examples}. We find that the major source of infelicitousness for positive propositions is propagated error from relation extraction\footnote{since we use automatic open relation extraction methods to replace human-annotated predicates, in order for scaling up the number of entailments inspected in our dataset.}; for negative propositions, the major source include the retrieved hyponym being semantically reasonable but pragmatically inappropriate, and the retrieved hyponym corresponding to an incorrect word sense for the respective positive.




\subsection{How Directionality Is Involved in BoOQA}
\label{supp:booqa_directionality}





In this section, we explain how BoOQA tests for directionality. Suppose in BoOQA we have a positive proposition \textcolor{red}{$A$} and its corresponding negative proposition \textcolor{red}{$B$}, which has WordNet hyponyms replaced into one of its spans. At evaluation, we retrieve from the context window a third proposition \textcolor{red}{$\it C$}, which we know is true and is under the same context as \textcolor{red}{$A$} and \textcolor{red}{$B$}. We have two measures of entailments, one is \textit{directional}, the other is \textit{symmetric}; both are perfect in their own right.

A perfect \textit{directional} measure is contemplated as a measure that would assign score 1 to all true entailments and 0 to all non-entailments; a perfect \textit{symmetric} measure would assign score 1 to all paraphrastic pairs, 0.5 to all semi-paraphrastic pairs (directional true entailments or directional non-entailments), and 0 to all unrelated pairs. This is an oversimplified hypothetical situation, where we ignore the non-categorical feature of predicate entailments; we note that this is only for the benefit of theoretical discussions.

There are 8 possible scenarios according to the genuine correctness of $\it \textcolor{red}{C} \vDash \textcolor{red}{A}$, $\it \textcolor{red}{A} \vDash \textcolor{red}{C}$, $\it \textcolor{red}{B} \vDash \textcolor{red}{C}$. We only care about these three edge out of the 6, because: 1) whether $\textcolor{red}{A}$ and $\textcolor{red}{B}$ entail each other is unimportant for determining whether $\textcolor{red}{A}$ or $\textcolor{red}{B}$ are true given $\it \textcolor{red}{C}$; 2) it must be the case the $\it \textcolor{red}{C} \nvDash \textcolor{red}{B}$ for the entry to be valid, otherwise the false proposition $\textcolor{red}{B}$ would be indorsed by the context. 

The goal of a method is to rank the positive higher than the negative in each scenario; when both methods rank the positive and the negative equally, the one that assigns lower confidence scores to them is considered the winner. This is because, for each proposition, there are multiple context propositions to refer to. Therefore, when evidence from the current context is insufficient for a judgement, being conservative harms less.

In 4 of the 8 scenarios 
, a perfect symmetric measure would do equally well as a perfect directional measure; however, in the other 4 scenarios %
, a perfect directional measure would prevail.%

This analysis illustrates how directional entailments are involved in our BoOQA evaluation. It also illustrates how we manage to test for a broad set of directional entailments by building a dataset with only hyponyms / hypernyms. The relation between $\textcolor{red}{A}$ and $\textcolor{red}{B}$ is hyponym / hypernym, however the entailments that we really test for are between the third proposition $\textcolor{red}{C}$ and each of $\textcolor{red}{A}$ / $\textcolor{red}{B}$, which can be all kinds of entailments.

\section{Computational Infrastructure and Model Footprint}
\label{supp:compute}

For our experiments, we use RTX 2080Ti GPUs with 11GB graphic memories. We set the computational limit for each job to be 168 GPU hours (equivalent to 7 days on 1 GPU). Our experiments are not CPU-intensive, but we note that for evaluation of entailment graphs on BoOQA,  $\sim 120$GB of memory is required for speedy processing with the context windows pre-loaded into memory.

S\&S fine-tuning on full LevyHolt dataset and the \textit{symmetric} subset typically take around 40 minutes to finish on these GPU; we run the fine-tuning with 100 different hyper-parameters for each experiment, so the total duration is approximately 60 GPU hours; S\&S fine-tuning on smaller subset typically take around 15 minutes for each run, and around 25 GPU hours for each experiment.

For evaluation on the BoOQA dataset, unsupervised LMs take $\sim$ 10 GPU hours; S\&S with base models take $\sim$ 40 GPU hours, S\&S with large models take $\sim$ 100 GPU hours. Evaluation of entailment graphs does not require GPU, which typically take 40-50 CPU hours with $\sim$120GB memory.

The $S\&S$ models have a similar \# of parameters to their underlying LMs, with only a linear layer added upon them; the rest of methods are either out-of-the-box LMs or discrete language resources.

\section{Details of BoOQA Baselines}
\label{supp:booqa_experiments}

In this section, we discuss some implementation details of BoOQA baselines, report $\it AUC_{norm}$ values for baselines on BoOQA dev sets in Table \ref{Tab:qaeval_supp_dev}, and discuss the presentation choices in Table \ref{Tab:qaeval}; we also present more baselines on the BoOQA test set.

For each method, we allow up to 3200 pieces of context evidence for each hypothesis. The only exception is $S\&S_{full}$ for $BoOQA_{ZH}$, where the speed is so slow that we had to cap this threshold to 90, hence the asterisk in Table \ref{Tab:qaeval}.

\begin{table}[t]
    \centering
    \begin{tabular}{|c|c|c|}
        \hline
        Dev & $BoOQA_{En}$ & $BoOQA_{Zh}$  \\\hline
        \multicolumn{3}{|l|}{LM-base} \\\hline
        $\it BERT_{tfidf}$ & 0.00 & 12.3 \\
        $\it BERT_{sent}$ & 0.3 & 9.8 \\
        $\it BERT_{rel}$ & 15.3 & 31.2 \\
        $\it RoBERTa_{rel}$ & 5.0 & 9.1 \\
        $S\&S_{Full}$ & 25.7 & 21.5 \\\hline
        \multicolumn{3}{|l|}{LM-large} \\\hline
        $\it BERT_{rel}$ & 16.9 & 26.9 \\
        $S\&S_{Full}$ & 24.9 & - \\\hline
        \multicolumn{3}{|l|}{Entailment Graphs} \\\hline
        EG BInc & 29.9 & 39.6 \\
        EG CNCE & 33.6 & - \\
        EG EGT2 & 25.9 & - \\\hline
    \end{tabular}
    \caption{Dev set results on English and Chinese BoOQA datasets for various baselines, values in \% of $\it AUC_{norm}$. All methods are out-of-the-box with no tuning; for Chinese we do not train $S\&S$ model with XLM-RoBERTa-large due to computational limits.}
    \label{Tab:qaeval_supp_dev}
\end{table}

\begin{table}[t]
    \centering
    \begin{tabular}{|c|c|c|}
        \hline
        Test & $BoOQA_{En}$ & $BoOQA_{Zh}$  \\\hline
        \multicolumn{3}{|l|}{LM-base} \\\hline
        $\it BERT_{tfidf}$ & 0.01 & 11.5 \\
        $\it BERT_{sent}$ & 0.08 & 10.8 \\
        $\it BERT_{rel}$ & 15.9 & 30.6 \\
        $\it RoBERTa_{rel}$ & 4.8 & 8.5 \\
        $S\&S_{Full}$ & 25.6 & 23.1 \\\hline
        \multicolumn{3}{|l|}{LM-large} \\\hline
        $\it BERT_{rel}$ & 17.7 & 26.0 \\
        $S\&S_{Full}$ & 24.8 & - \\\hline
    \end{tabular}
    \caption{More test set baseline results on English and Chinese BoOQA dataset, values in \% of $\it AUC_{norm}$. All methods are out-of-the-box with no tuning.}
    \label{Tab:qaeval_supp_test}
\end{table}

\paragraph{$\bf LM_{unsupervised}$} Firstly is the context retrieval approaches for unsupervised LM methods. We experimented with three variants: 

\begin{itemize}[itemsep=0pt]
    \item $\it BERT_{tfidf}$: for each hypothesis, retrieves the top 5 most relevant articles by TF-IDF measure following \citet{chen_reading_2017};
    \item $\it BERT_{sent}$: for each hypothesis, retrieves the \textit{host-sentences} of all relations involving the same arguments as the queried hypothesis;
    \item $\it BERT_{rel}$: for each hypothesis, retrieves the \textit{relations} involving the same arguments as the queried hypothesis (in short sentences);
\end{itemize}

Among them, $\it BERT_{rel}$ outperforms the other two, likely because it uses more focused context input of concatenated relations rather than the more noisy raw news sentences. 

Between BERT and RoBERTa LMs, Although RoBERTa is shown \cite{schmitt_language_2021} to be good for prompting with LevyHolt, BERT has a clear advantage for unsupervised cosine similarity. Between BERT-base and BERT-large, the more computationally expensive BERT-large model does not show a clear advantage, so we stick to the base version. Therefore, on test set we report $\it BERT_{rel}$ results with BERT-base as $\it LM_{unsupervised}$ in Table \ref{Tab:qaeval}, and leave the rest to Table \ref{Tab:qaeval_supp_test}.


\paragraph{$\bf S\&S$ \textbf{Models}} For $S\&S$ methods we additionally report baselines trained on full LevyHolt with RoBERTa-large. Interestingly $S\&S_{Full} large$ does not outperform $S\&S_{Full} base$ on BoOQA, therefore in Table \ref{Tab:qaeval}, we stick with RoBERTa-base.

\begin{figure}[t]
    \centering
    \includegraphics[scale=0.45]{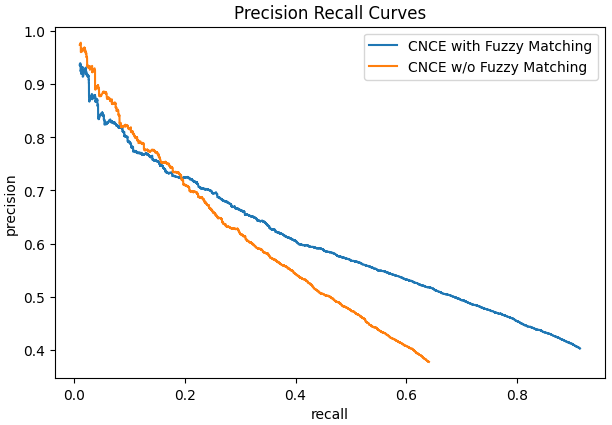}
    \caption{Precision-recall curves of CNCE entailment graphs on BoOQA dev set, w. / w.o. fuzzy matching.}
    \label{Fig:fuzzy_matching}
\end{figure}

\paragraph{EG} Here for EG methods, we include a discussion on fuzzy matching and sparsity. Entailment graphs represent predicates in the form like ``(go.1, go.to.2)'', where the subject and object semantic roles are indicated by the two parts separated by a comma. In practice, multiple semantic role assignments can be given to one predicate in natural language form; so with fuzzy matching, we ground the natural language predicates in propositions to entailment graph nodes regardless of the semantic role assignments they may be associated to.

The fuzzy matching also brings the entailment graph methods to the same page as LM-driven methods: looking only at the natural language forms. As shown in Figure \ref{Fig:fuzzy_matching}, without fuzzy matching the recall hit a wall at 64.1\%, where sparsity is at full display. With fuzzy matching, much more entries are assigned some score, bringing the right boundary of recall to 91.4\%; however, on one hand this harms the precision for top-notch edges, on the other hand the precision still drops rapidly with increasing recall, meaning that beyond the most confident entailment edges, the quality of entailment graph edges deteriorates quickly. That is to say, sparsity still hampers the performance of entailment graphs.




\section{The Directional Subset with \citet{chen_entailment_2022}}
\label{supp:egt2}

\begin{table}[t]
    \centering
    \begin{tabular}{|c|c|}
    \hline
        Method & $\it AUC_{50\%}$ \\\hline
        BInc \cite{szpektor_learning_2008} & 54.7 \\
        Weeds \cite{weeds_general_2003} & 55.2 \\
        CNCE \cite{hosseini_open-domain_2021} & 55.9 \\
        EGT2 \cite{chen_entailment_2022} & 63.3 \\\hline
    \end{tabular}
    \caption{Performance of various entailment graphs on the LevyHolt \textit{directional} subset, values in \%. }
    \label{Tab:egt2_dir_subset}
\end{table}

Contemporary to us, \citet{chen_entailment_2022} presented a new entailment score for building entailment graphs, where they first fine-tune a DeBERTa model on MNLI, then re-fine-tune it with pairs of short propositions composed of (subject, predicate, object) triples, using PPDB annotations for supervision, in order for domain transfer. They reported results of their EGT2 graph on a version of \textit{directional} subset of LevyHolt. However, we have found their \textit{directional} subset to be different from that first presented by \citet{holt_probabilistic_2019}. 

We re-evaluate the EGT2 graph on the \citet{holt_probabilistic_2019} original version of \textit{directional} subset and list the results in Table \ref{Tab:egt2_dir_subset}, which are presented in $\it AUC_{50\%}$. Consistent with the results reported in \citet{chen_entailment_2022}, the EGT2 graph still outperforms the rest of the graphs, albeit not by as much. This introduces a slight inconsistency between results on the LevyHolt \textit{directional} subset and the BoOQA dataset; we trust the larger BoOQA dataset and speculatively argue that the instability of the small LevyHolt \textit{directional} test set and some transferrable spurious correlations in MNLI related to the behavior of human annotators may be to blame.

One last remark is, there is a slight difference in the calculation of $\it AUC$ between \citet{hosseini_learning_2018}/\citet{chen_entailment_2022}, and \citet{schmitt_language_2021}, in terms of left boundaries of recall, where the calculations by \citet{schmitt_language_2021} yield higher numbers. Since the calculation by \citet{schmitt_language_2021} is compatible with all methods, we use the this calculation throughout.

\end{CJK*}
\end{document}